\documentclass[lettersize,journal]{IEEEtran}
\usepackage{CJKutf8}

\usepackage{amsmath,amsfonts}
\usepackage{algorithmic}
\usepackage{algorithm}
\usepackage{array}
\usepackage[caption=false,font=normalsize,labelfont=sf,textfont=sf]{subfig}
\usepackage{textcomp}
\usepackage{stfloats}
\usepackage{url}
\usepackage{verbatim}
\usepackage{graphicx}
\usepackage{cite}
\usepackage{hyperref}
\usepackage{multirow}
\usepackage{enumitem}
\usepackage{color,soul}
\usepackage{xcolor}
\setlist[enumerate]{leftmargin=*}
\begin{document}

\begin{CJK}{UTF8}{gbsn}

\title{Going the Extra Mile in Face Image Quality Assessment: A Novel Database and Model}

\author{Shaolin Su$^{a,c}$, Hanhe Lin$^{b,c}$, Vlad Hosu$^c$, Oliver Wiedemann$^c$, Jinqiu Sun$^a$, Yu Zhu$^a$,\\
Hantao Liu$^d$, Yanning Zhang$^a$ and Dietmar~Saupe$^c$
        % <-this % stops a space
\thanks{Funded by the Deutsche Forschungsgemeinschaft (DFG, German Research Foundation) -- Project-ID 251654672 -- TRR 161 (Project A05). Corresponding author: Hanhe~Lin, E-mail: hlin001@dundee.ac.uk}% <-this % stops a space
\thanks{$^a$School of Computer Science and Engineering, Northwestern Polytechnical University, China.}
\thanks{$^b$School of Science and Engineering, University of Dundee, DD1 4HN Dundee, United Kingdom.}
\thanks{$^c$Department of Computer and Information Science, University of Konstanz, 78464 Konstanz, Germany.}
\thanks{$^d$School of Computer Science and Informatics, Cardiff University, CF24 4AG Cardiff, United Kingdom.}
}

\markboth{IEEE Transactions on Multimedia,~Vol.~xx, No.~x, August~2023}%
{Su \MakeLowercase{\textit{et al.}}: Going the Extra Mile in Face Image Quality Assessment: A Novel Database and Model}


\maketitle

\begin{abstract}

An accurate computational model for image quality assessment (IQA) benefits many vision applications, such as image filtering, image processing, and image generation. Although the study of face images is an important subfield in computer vision research, the lack of face IQA data and models limits the precision of current IQA metrics on face image processing tasks such as face superresolution, face enhancement, and face editing. To narrow this gap, in this paper, we first introduce the largest annotated IQA database developed to date, which contains 20,000 human faces -- an order of magnitude larger than all existing rated datasets of faces -- of diverse individuals in highly varied circumstances. Based on the database, we further propose a novel deep learning model to accurately predict face image quality, which, for the first time, explores the use of generative priors for IQA. By taking advantage of rich statistics encoded in well pretrained off-the-shelf generative models, we obtain generative prior information and use it as latent references to facilitate blind IQA. The experimental results demonstrate both the value of the proposed dataset for face IQA and the superior performance of the proposed model.
\end{abstract}

\begin{IEEEkeywords}
Image quality assessment, face quality, subjective study, GAN, generative priors
\end{IEEEkeywords}

\section{Introduction}

\IEEEPARstart{T} he computer vision research on human faces includes the key area of multimedia processing. 
Since the human visual system (HVS) is especially sensitive to human faces \cite{perrett1987visual, o1997areal} in media content, dedicated processing tasks such as face superresolution, face enhancement, face generation and face editing have garnered growing interest over the past few decades. Although quality control in face image processing applications is a crucial factor determining user experience, the lack of face image quality metrics in the current research limits the precise measurement of these face-specific applications. Recently, blind image quality assessment (BIQA) approaches applied to broad-domain images have significantly improved; however, it is still unclear whether the methods are directly applicable to the face domain due to the following two factors:
First, because of the specific processing mechanism dedicated to faces in the HVS \cite{perrett1987visual,ro2001changing,theeuwes2006faces}, the perceptual representation and mapping pattern to face quality might be different from those to generically categorized images. Therefore, learning a dedicated face quality metric not only improves the quality prediction accuracy but also assists in understanding the perceptual mechanism of the HVS for human faces. Second, as existing IQA databases collect images of mostly generic categories, face image data are less often included. For example, only approximately 2\% and 10\% of images contain faces in two of the largest in-the-wild IQA datasets, KonIQ-10k \cite{hosu2020koniq} and SPAQ \cite{2020Perceptual}, respectively. As a result, either the image content shift or sample amount limits the ability of existing IQA models to draw correct quality mappings to face data.

Consequently, there is a need for IQA datasets that contain more subjectively rated face images to facilitate face IQA and processing. In this paper, we therefore introduce such a large-scale quality-annotated dataset and expect several applications to benefit from generic face image quality assessment (GFIQA). Examples of potential applications are as follows: 1. To improve the performance of face recognition, face images with quality scores below a predetermined threshold can be excluded during the acquisition phase, hence reducing the error rate of face recognition systems.
2. To improve general IQA model predictions, as the HVS is extremely sensitive to faces, the visual quality of face regions might be more critical in the perception process of the whole image; Therefore, an accurate face IQA metric could be advantageous for the general IQA task. 
3. Other practical usages include album selection and optimization. When importing images to a photo album, face image quality can be used as a standard to determine acceptance or rejection.

Note that GFIQA differs from the definition of face image quality assessment in the biometrics community~\cite{grother2020ongoing,schlett2021face}, where quality is a form of utility for biometric systems such as identification of a face image. 
Recently, \cite{tian2022generalized} also proposed a method to assess the visual quality of face images; however, they focused on GAN-generated face images, which address the quality assessment related to image synthesis models.
Different from the above, for GFIQA, we aim to create predictive models (metrics) for in-the-wild face image quality assessment, where the quality relates to the degradation factors existing in the real world. The factors include the degradation introduced by an imaging system during capturing, processing, storage, compression, and display of face images~\cite{liu2019pre,sun2022graphiqa}.

To compute accurate estimates of generic face IQA, we further proposed a novel model to fulfill the task. The recent successful use of deep generative priors in many image restoration and editing tasks \cite{xia2021gan} has inspired us to explore the effectiveness of this powerful guidance in the field of IQA.
In contrast to previous models using a vanilla encoder~\cite{yang2020ttl,yang2020blind,ou2021novel}, we are the first to exploit deep generative priors in an image quality prediction model and develop an effective framework for utilizing these powerful priors in IQA. 
% rich statistics of natural images are encoded, which could be utilized as latent references for the blind IQA task. The combination of distorted and latent reference features therefore allows for more accurate quality prediction results.
Rich statistics of natural images are encoded in pretrained generative models; by extracting intermediate generative features, we can utilize them as latent references corresponding to the distorted target images. The combination of distorted and latent reference features therefore facilitates the blind IQA task and allows for more accurate quality prediction results. Note that different from previous GAN-based IQA models \cite{lin2018hallucinated, ko2020quality} that train their generators from scratch, the proposed model directly makes use of off-the-shelf GAN models to extract the prior information. The framework not only avoids the cumbersome training procedure of the generative models but also leverages the well pretrained GAN models on large-scale image data as an approximation to the natural image manifold, thus possessing more stability toward solving the challenging in-the-wild IQA problem.

The main contributions of this paper are as follows:
\begin{enumerate}
    \item We created the largest IQA database of human faces in-the-wild, which is called the \textit{Generic face image quality assessment 20k database} (GFIQA-20k). We collected 20,000 face images and ensured the diversity of the individuals, who are depicted in highly varied circumstances. 
    We also validated the reliability of the collected dataset with gold-standard questions and self-consistency tests.

    \item We proposed a novel quality prediction model that for the first time employs deep generative priors to facilitate the BIQA task. Using the rich statistics encoded in pretrained generative models, we obtain prior preserved images and use them as latent references to improve the IQA prediction accuracy. 
    
    \item The experimental results verified both the usefulness of the proposed dataset in evaluating face image quality and the effectiveness of the proposed model in achieving accurate predictions. The database and code will be made available at \url{http://database.mmsp-kn.de/gfiqa-20k-database.html}.
\end{enumerate}

\section{Related Works}

\subsection{Quality Assessment of Face Images}

There are two main research areas that address the quality of face images. The first and most developed field stems from the biometrics community and aims at assessing face image quality for face recognition systems. This is most often referred to as face image quality assessment (FIQA). The second is GFIQA and relates to general image quality assessment dealing with perceptual image degradation. An in-depth discussion about the differences between the two fields has been presented by Schlett \textit{et al.}~\cite{schlett2021face}.

FIQA has attracted increased attention in the face recognition community \cite{grother2020ongoing,schlett2021face}. Earlier works proposed measuring the quality of a face image in terms of its similarity to its reference face image with respect to multiple factors such as pose, expression, illumination, and occlusion. For example, Sellahewa and Jassim~\cite{sellahewa2010image} measure image quality in terms of luminance distortion by comparing a face input image to a known reference image. However, such approaches are difficult to apply since they must consider every possible factor individually, and reference face images may not be available in an unconstrained environment. 

In contrast, learning-based approaches, where the target face quality is defined in some manner to be indicative of face recognition performance, are more favorable. These learning-based approaches can be grouped according to the way the ground truth quality values are labeled. In most approaches, the ground truth quality values are determined computationally. For instance, Bharadwaj \textit{et al.}~\cite{bharadwaj2013can} assigned qualities to face images by using two commercial off-the-shelf face recognition systems, where a face image is given a good quality value if it matches well. 
Chen \textit{et al.}~\cite{chen2014face} assumed that the face images in dataset $A$ have better quality than those in dataset $B$ for a face recognition method if the recognition performance of this method on $A$ is better than that on $B$. 
Although there exists work that labels face quality manually, e.g., in binary classes (good or bad)~\cite{zhao2019face}, Best-Rowden and Jain~\cite{best2018learning} conducted the first subjective face quality assessment study. By conducting a study on a small set of pairwise comparisons of 13,233 face images taken from \cite{LFWTech},
the quality ratings of all images were inferred using matrix completion. 

While FIQA is evolving significantly, no studies on GFIQA exist. To the best of our knowledge, except for the small number of face images present in existing IQA datasets, our work is the first study of this kind dedicated exclusively to face images. 

\subsection{Generative Priors}

With the rapid development of generative models, GANs have become capable of effectively learning the natural image manifold and synthesizing high-resolution images with pleasant visual quality \cite{karras2018progressive, brock2018large, karras2019style, karras2020analyzing}. With pretrained GAN models, the well-learned image manifold can be further explored to promote diverse image manipulation and restoration tasks \cite{wang2018high, abdal2019image2stylegan, abdal2020image2stylegan++, pan2020exploiting, wang2021towards, wu2021towards}, referred to as generative priors. To utilize the rich information encoded in generative models, target images are first mapped back to the intermediate features or latent space of pretrained GANs \cite{xia2021gan}, and image manipulation or restoration tasks are then facilitated by feeding forward the inverted features or codes to generators.

There are typically two approaches to invert GANs and utilize the priors, optimization-based and learning-based. Optimization-based methods optimize the input code of the generator by minimizing the reconstruction error of the target image. By manipulating latent codes or modifying objective functions, image manipulation or restoration results can be obtained. Image2StyleGAN \cite{abdal2019image2stylegan} and Image2StyleGAN++ \cite{abdal2020image2stylegan++} optimized latent codes in the $\mathcal{W}$ space of StyleGAN \cite{karras2019style} and the $\mathcal{W}^{+}$ space of StyleGAN2 \cite{karras2020analyzing} and achieved image inpainting, morphing and style mixing results. mGANPrior \cite{gu2020image} optimized multiple latent codes and adaptively fused them to achieve various image restoration results, including colorization, superresolution, and denoising. Noticing the distribution gap between the training and testing data, DGP \cite{pan2020exploiting} proposed a method to fine-tune generator parameters on-the-fly to adapt the target images while maintaining the statistics of the GAN learned priors. Although they require no training procedures, optimization-based methods are usually time-consuming due to the large iteration numbers needed for each instance image.

Learning-based approaches train an extra encoder to map an image to its latent code. By modifying encoder architectures, various image manipulation and reconstruction results can be achieved. pSp \cite{richardson2021encoding} trained a multistage encoder to generate a series of style codes for StyleGAN2 \cite{karras2020analyzing} and handles various facial image translation tasks, including conditional image synthesis, facial frontalization, inpainting, \emph{etc.}. GLEAN \cite{chan2021glean} proposed an encoder-generator-decoder design to fulfill the large-factor image superresolution task. GFP-GAN \cite{wang2021towards} and GPEN \cite{yang2021gan} fused target image features with generative prior features to restore real-world degraded face images. \cite{wu2021towards} warped and modulated generative prior features to achieve controllable image colorization results. Unlike optimization-based approaches, learning-based approaches obtain image restoration results by performing only one feed-forward pass at test time. However, extra data are usually needed to train these models.

In this paper, we investigate for the first time the potential application of generative priors to the task of IQA. Specifically, we employ rich statistics encoded by StyleGAN2 as latent references from the pristine image manifold to facilitate the solving of the blind IQA problem. To utilize generative priors efficiently and effectively, we propose a method to train a multistage encoder and take advantage of multilevel attributes controlled by the style codes to obtain the generative statistics. The proposed approach avoids both expensive optimization procedures and extra data training and shows its superiority in solving the objective face IQA problem.

\section{The Creation of the GFIQA-20k Database}

\subsection{Face Image Collection}

\textcolor{black}{
%Quality Control Editor: Please ensure that the edit maintains the intended meaning.
Our goal is to build an ecologically valid face IQA database that includes a wide range of user-generated face images representing those in the real world and thus train a model that is of ecological validity.}

To establish the dataset, we first collect face images from YFCC100M~\cite{thomee:2016}, a massive public multimedia database, to ensure diverse in-the-wild image qualities. We randomly selected and downloaded one million images, from which face images were extracted as follows. For a given image, we applied the MTCNN model \cite{zhang2016joint} to detect faces and their corresponding key points, where the minimum size parameter of the face to detect was set as $400$ pixels. Next, we aligned the image for each detected face according to the positions of the detected left and right eyes. The central point of a detected face was estimated to be the midpoint between the left and right eyes. Next, the detected face image was cropped such that both the width and height of the crop were equal to four times the distance between the left and right eyes. Finally, the crop was rescaled to $512 \times 512$ pixels. Using this procedure, we collected $86,026$ face images in the wild. 

In practice, the face detection model cannot always guarantee the absence of incorrect face detection instances, such as false-positives (not human faces), or inaccurate key points. In light of this, we manually checked and removed incorrectly detected faces. This step reduced the number of samples to $53,058$.

\begin{figure}[tb]
\centering
\includegraphics[width=0.49\textwidth]{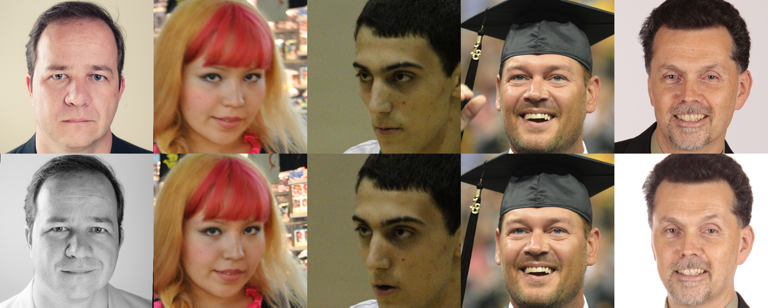}
\caption{Example of duplicate image pairs or image pairs with the same identity, where the top images are face images in GFIQA-20k, and the bottom images are images excluded from GFIQA-20k using our sampling strategy.}
\label{fig:duplicate}
\end{figure}

In the final step, to ensure identity diversity, we removed duplicate identities from the collected face images. We used the FaceNet model~\cite{schroff2015facenet} to extract the 512-dimensional deep features from the face images, which have been demonstrated to be effective in clustering face images into groups of people with the same identity. 
We next applied $k$-means clustering on the deep features to partition the $53,058$ images into 20,000 clusters. In this case, images in the same cluster will include duplicate images or face images with the same identity, as shown in Fig.~\ref{fig:duplicate}. In each cluster, an image is randomly selected as a representative. With this step, the number of face images decreased to 20,000, which were the face images included in the GFIQA-20k. With our sampling strategy, duplicated face images or images with the same identity were removed, which guarantees identity diversity.

\begin{figure}[tb]
\centering
\includegraphics[width=0.4\textwidth]{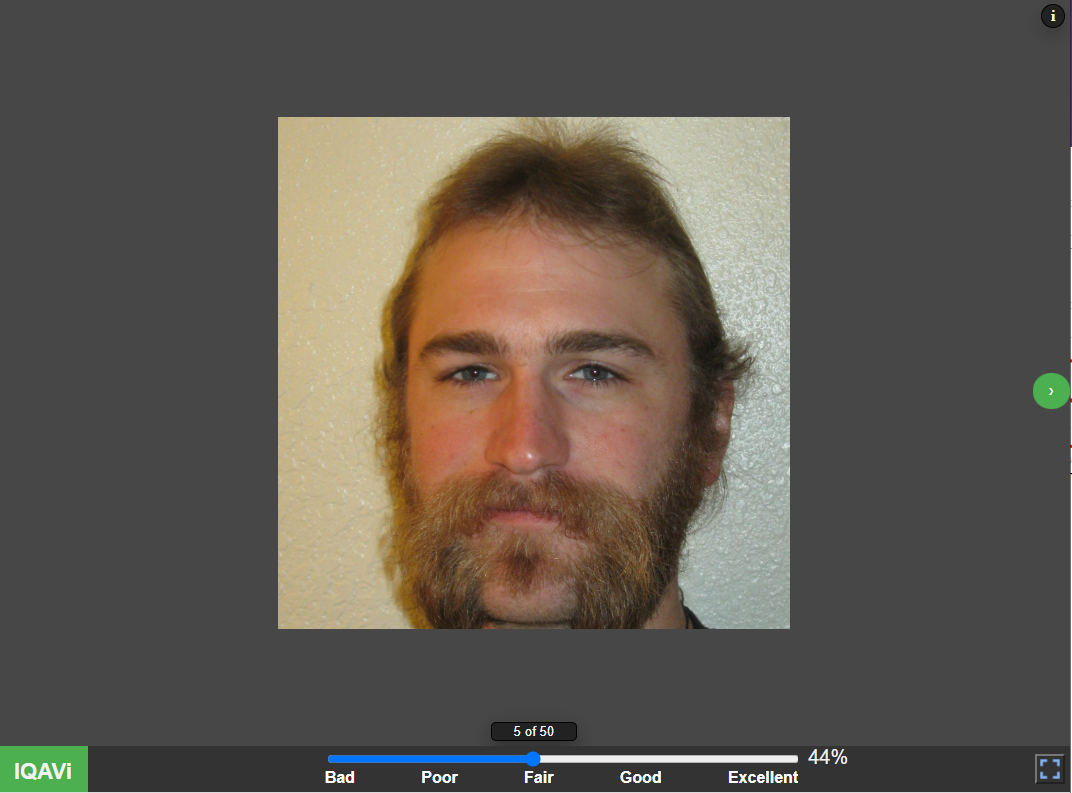}
\caption{UI for the subjective generic face IQA study. Each time participants were presented an image within a batch, they dragged a slider below the face to rate its visual quality on a scale ranging from Bad (1\%) to Excellent (100\%).}
\label{fig:ui}
\end{figure}

\subsection{Subjective Face Image Quality Assessment}

In the following, we performed a large-scale subjective study to assess the visual quality of 20,000 face images. The 20,000 images were randomly divided into 500 batches, where each batch initially contained 40 images. To better monitor and analyze participants' performance, two reliability mechanisms were used. One involved adding gold-standard or test data for which the correct answers are already known \cite{le2010ensuring}. The other involved utilizing a consistency test by posing the same question multiple times \cite{hossfeld2011quantification}. In our study, we manually selected 100 high-quality and 100 low-quality face images as gold-standard images. Five images were randomly sampled (with replacement) from the 200 images and added to each batch. Moreover, five of the 40 study images were presented twice in each batch. Eventually, each batch contained 50 images to be rated.

% from supp

\begin{figure*}[ht]
\centering
\begin{minipage}{0.2\linewidth}
\centering
\centerline{\includegraphics[width=\textwidth]{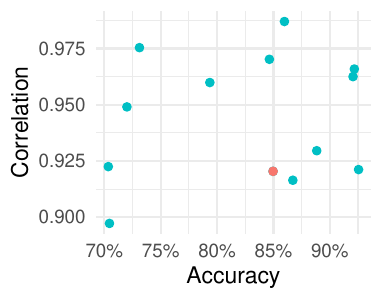}}
\text{(a)}
\end{minipage}
\begin{minipage}{0.32\linewidth}
\centering
\centerline{\includegraphics[width=\textwidth]{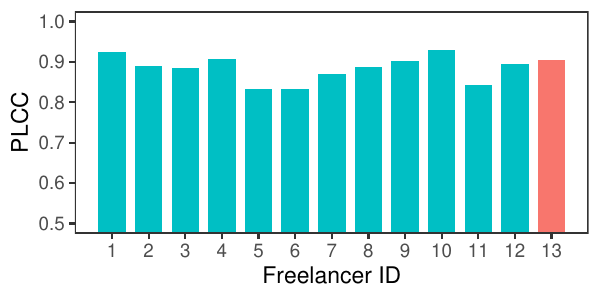}}
\text{(b)}
\end{minipage}
\begin{minipage}{0.32\linewidth}
\centering
\centerline{\includegraphics[width=\textwidth]{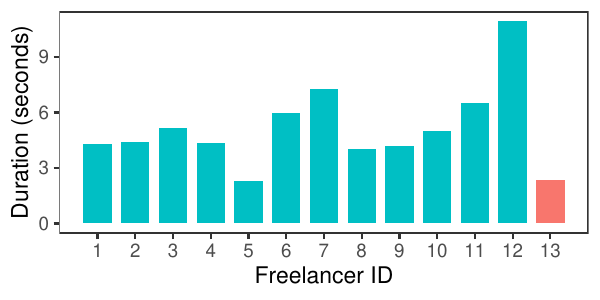}}
\text{(c)}
\end{minipage}
\caption{\textcolor{black}{Analysis of the subjective study. All participants submitted ratings for all images, except for the one shown in red, who submitted 153 batches (6,120 study images). (a) Reliability analysis of the subjective ratings. (b) Correlation between ratings of each individual freelancer and MOS. The PLCC values range from 0.832 to 0.928, all above the acceptance threshold of 0.75 PLCC, as recommended by \cite{RecommendationITUT913}. (c) The average duration (second) of rating an image for each subject in our study. The average duration is 5.25 seconds.}}
\label{fig:study_analysis}
\end{figure*}

% Instruction
Before carrying out the study, participants were first presented with a page of instructions containing four sections. In the first section, the definition of technical image quality was introduced. The hardware requirements and detailed study steps are explained in the second and third sections, respectively. In the final section, in addition to providing examples with different quality scales, we also provided some examples to differentiate between technical face image quality and face attractiveness.
The user interface for our subjective face IQA study is illustrated in Fig.~\ref{fig:ui}.

The standard 5-point absolute category rating (ACR) scale, i.e., Bad, Poor, Fair, Good, and Excellent, is used for subjective rating. Participants are presented with a batch of face images one at a time.
Each time participants dragged a slider below the face image to rate its visual quality on a scale ranging from bad (1\%) to excellent (100\%). As participants were required to drag a slider on a scale, which we linearly mapped to the interval [0.01, 1]. To be more specific, let $x$ be the original 5-point ACR, and the mapped score is $y=(x-1)/4 \times 0.99+0.01$. As a result, the mapped 5-point ACR on the slider is Bad -- 1\%, Poor -- 25.75\%, Fair -- 50.5\%, Good -- 75.25\%, and Excellent -- 100\%. 

% Training session
To guide the freelancers on using the interface, we provided a training session for them. It contained 60 face images with given answers collected from the KonIQ-10k IQA database~\cite{hosu2020koniq}. After giving a quality rating for an image, freelancers could click a button to proceed to the next image. However, if the assessment result was incorrect, they were informed of the incorrect assessment, and a range for the slider position was suggested. Freelancers could only proceed after having moved the slider into the suggested correct range. \textcolor{black}{In the process of study, the duration of rating an image was recorded for further analysis.}

A total of 13 freelancers were hired to participate in this study, 7 of whom are visual arts professionals such as designers, graphics artists, and photographers. More importantly, they all achieved excellent performance in a previous IQA contest of ours (not published), which demonstrated their expertise in IQA. One freelancer quit the study after submitting 153 batches, while the remaining freelancers completed the entire study.

\subsection{Subjective study analysis}

\begin{figure*}[t]
\centering
\begin{minipage}{0.69\linewidth}
\centering
\centerline{\includegraphics[width=1.0\textwidth]{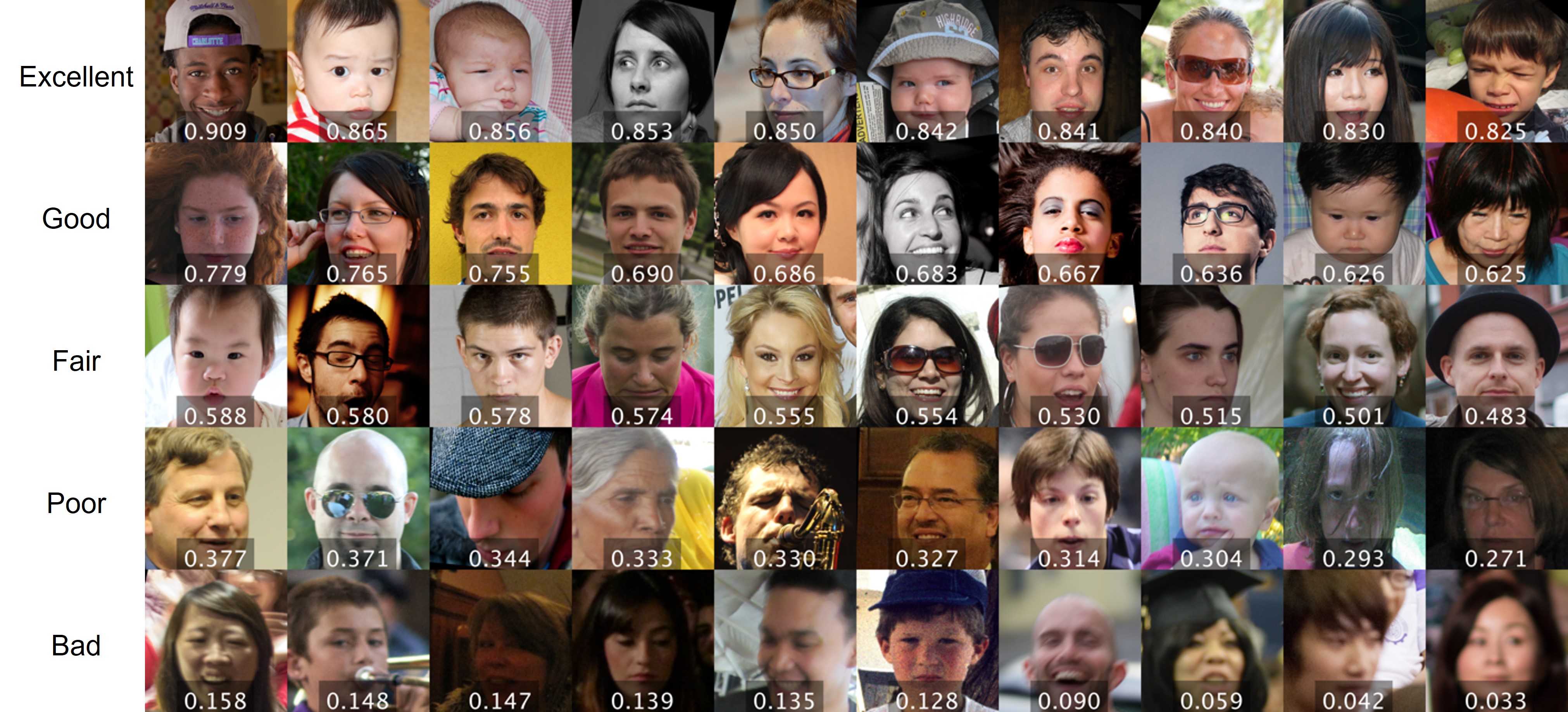}}
\text{(a)}
\end{minipage}
\begin{minipage}{0.30\linewidth}
\centering
\centerline{\includegraphics[width=\textwidth]{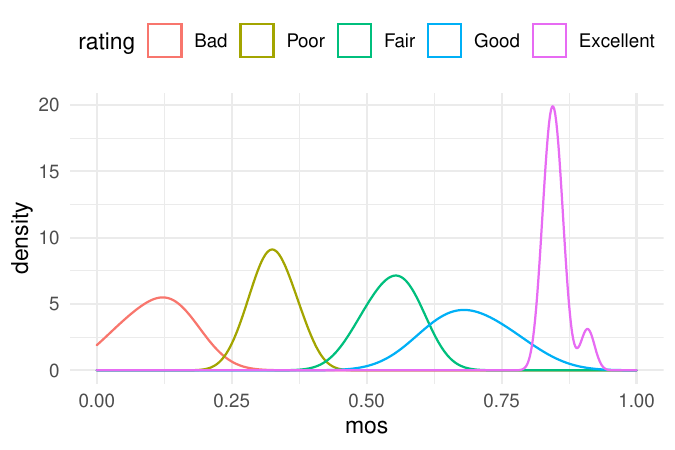}}
\text{(b)}
\end{minipage}
\caption{
\textcolor{black}{ We divide the images into the categories of Excellent ((0.8,1]), Good
((0.6,0.8]), Fair ((0.4,0.6]), Poor ((0.2,0.4]), and Bad ([0,0.2]), where 10 images are randomly sampled from each category. (a) Sampled face images with their corresponding MOS (white digits) and categories in the GFIQA-20k dataset. (b) MOS distribution according to different categories.
}}
\label{fig:face_example}
\end{figure*}

\begin{figure*}[t]
\centering
\begin{minipage}[t]{0.24\linewidth}
   \includegraphics[width=1\textwidth]{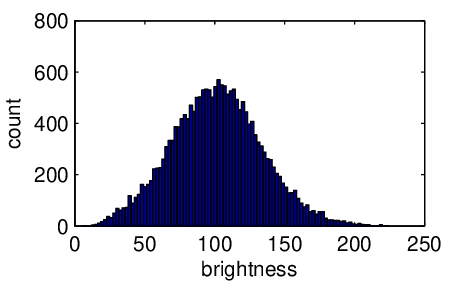}
   \centering
\end{minipage}
\begin{minipage}[t]{0.24\linewidth}
   \includegraphics[width=1\textwidth]{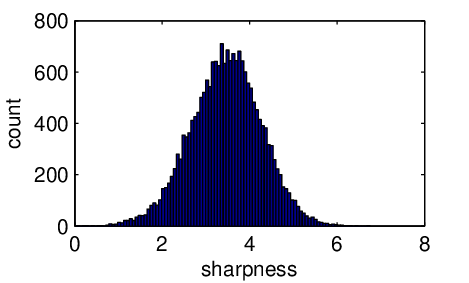}
   \centering
\end{minipage}
\begin{minipage}[t]{0.24\linewidth}
   \includegraphics[width=1\textwidth]{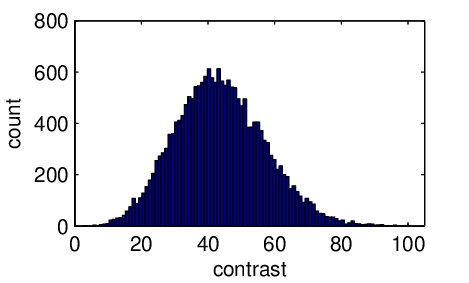}
   \centering
\end{minipage}
\begin{minipage}[t]{0.24\linewidth}
   \includegraphics[width=1\textwidth]{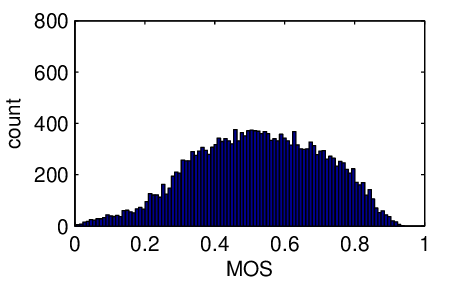}
   \centering
\end{minipage}
\caption{\textcolor{black}{We show the distributions of the dataset in regard to several aspects, including lighting, sharpness, contrast, and MOS distribution, to visualize the collected data attributes.}}
\label{fig:distributions}
\end{figure*}

We determined the reliability of the freelancers by measuring their accuracy on gold-standard test images and correlation on a self-consistency test. \textcolor{black}{Before conducting the analysis, min-max normalization was applied for the rating of each subject.} For a gold-standard test image, a freelancer’s answer is counted as correct if his answer falls in the range of 1\% to 35\% when the image is labeled as low quality or in 65\% to 100\% when the image is labeled as high quality. For the self-consistency test, we used Spearman's rank correlation coefficient (SRCC) to calculate the reliability. 
The statistics of the freelancer reliability analysis are shown in Fig.~\ref{fig:study_analysis}(a).
In Fig.~\ref{fig:study_analysis}(a), we plot the accuracy achieved on gold-standard images on the x-axis and the self-consistency on repeatedly presented images, expressed as the SRCC between the two scores provided for all images, on the y-axis. All participants achieved excellent self-consistency (mostly over 0.9 SRCC) and maintained a high level of accuracy relative to the gold-standard ratings (over 70\% accuracy).

In addition to the reliability analysis, we report the Pearson linear correlation coefficient (PLCC) between individual ratings and MOS in Fig.~\ref{fig:study_analysis}(b). 
The ratings of each freelancer are highly correlated with the MOS, ranging from 0.832 to 0.928, all above the acceptance threshold of 0.75 PLCC, as recommended by \cite{RecommendationITUT913}. The results demonstrate that all participants achieve high reliability regarding the subjective ratings, thus guaranteeing the reliability of the constructed dataset. 

\textcolor{black}{Fig.~\ref{fig:study_analysis}(c) shows the average duration of rating an image for each subject. The duration for each subject varies from 2.29 sec to 10.92 sec, with an average duration of 5.25 sec. }

In the final step, to further improve the rater agreement, we perform outlier detection and removal procedures. We screen the ratings based on the assumption that the ratings provided by reliable participants lie in an interval around the mean of all the ratings of an image. To be more specific, the interval's length is twice the standard deviation of all ratings from an image; ratings outside the interval are removed, and the rest yield the mean opinion scores (MOSs). For each image, an average of 12 ACR ratings are obtained.

\subsection{\textcolor{black}{Database overview}}

\textcolor{black}{
Finally, the collected MOSs with the corresponding 20,000 face images form the proposed GFIQA-20k dataset. In this subsection, we provide an overview and some analyses of the established dataset.}

\textcolor{black}{
The database contains images with MOSs in the range of 0.005 to 0.941.
In Fig.~\ref{fig:face_example}, we show image samples of different categories (Excellent, Good, Fair, Poor, and Bad) from the GFIQA-20k dataset. We sample 10 images from each category and plot their MOS distributions in Fig.~\ref{fig:face_example} (b). As seen, the collected face images cover a diverse perceptual quality range, while images belonging to different categories are distinguishable from each other in the MOS distributions.}
\textcolor{black}{
In Fig.~\ref{fig:distributions}, we further show the distribution of the data in several dimensions, including brightness, sharpness, contrast and MOS distribution. In our calculation, brightness is estimated by the mean grayscale value, sharpness is calculated by taking the log of the image gradient magnitudes, and contrast can be calculated by $contrast(I)=std(I)/kurtosis(I)^{\frac{1}{4}}$, where $std(I)$ and $kurtosis(I)$ are the standard deviation and kurtosis of the image signal, respectively. It can be seen that the collected data is diverse in terms of different measurements, thus demonstrating the richness of the dataset.}

\begin{figure}[t]
\centering
\begin{minipage}{0.23\linewidth}
\centering
\centerline{\includegraphics[width=\textwidth]{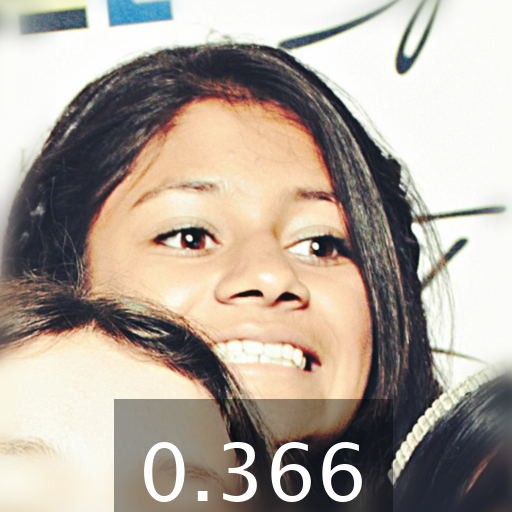}}
\text{(a)}
\end{minipage}
\begin{minipage}{0.23\linewidth}
\centering
\centerline{\includegraphics[width=\textwidth]{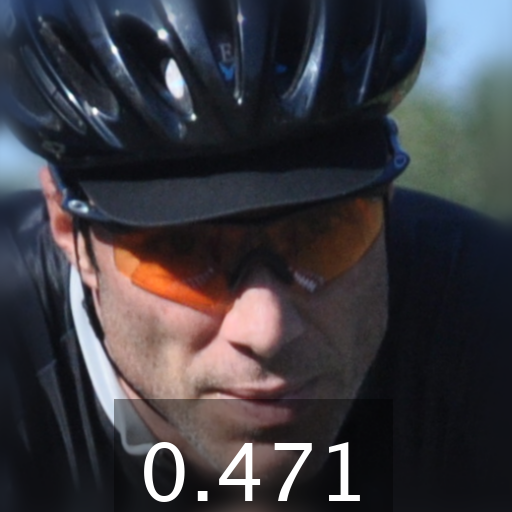}}
\text{(b)}
\end{minipage}
\begin{minipage}{0.23\linewidth}
\centering
\centerline{\includegraphics[width=\textwidth]{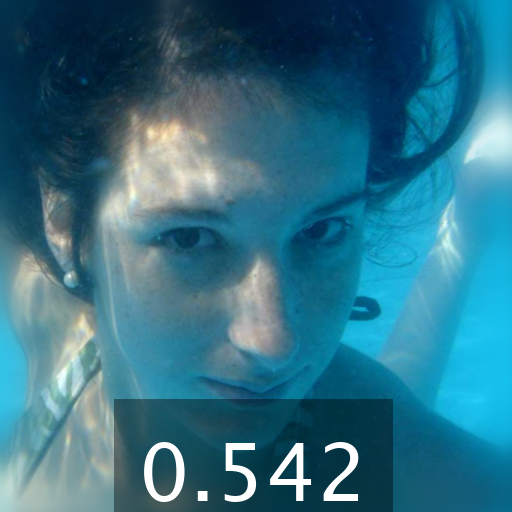}}
\text{(c)}
\end{minipage}
\begin{minipage}{0.23\linewidth}
\centering
\centerline{\includegraphics[width=\textwidth]{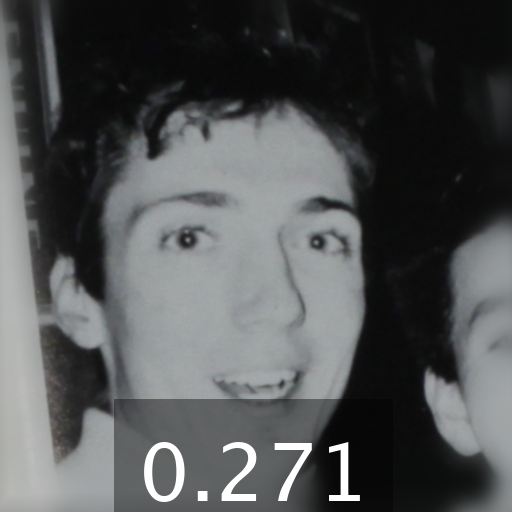}}
\text{(d)}
\end{minipage}
\caption{\textcolor{black}{We show some types of distortions that particularly affect face image quality in addition to common in-the-wild distortions. The distortions include occlusion/shadows on faces ((a) and (b)), underwater faces (c) and faces from old photos (d). }}
\label{fig:other_dist}
\end{figure}

\textcolor{black}{
Although the proposed dataset collects data from the real world and thus contains distortions similar to those in the in-the-wild IQA dataset, we find that there exist other distortions that particularly affect face quality. We show some examples as well as their MOSs in Fig.~\ref{fig:other_dist}; the distortions include occlusion/shadow on face regions, underwater faces and faces from old photos \textit{etc.}. This indicates that the HVS perceives face quality differently from generic image quality and the need for constructing a specific face IQA dataset.}

\section{Face Image Quality Assessment with Generative Priors}

\begin{figure*}[t]
\centering
\includegraphics[width=1\textwidth]{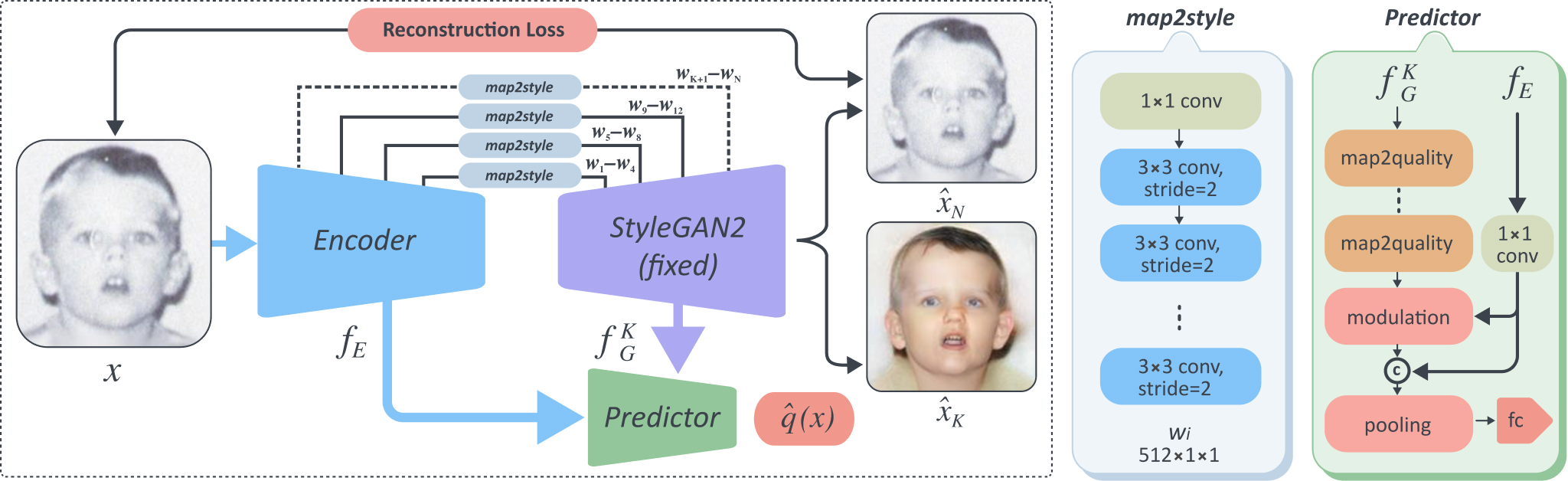}
\caption{The proposed face IQA model utilizing generative priors. Our framework consists of the following three parts: an encoder to both invert the target image and extract distortion features, a generator to produce latent reference features in a pretrained GAN space, and a predictor to make quality estimations by refining and fusing target image features and latent reference features.
}
\label{fig:arch}
\end{figure*}

In this section, we provide a detailed description of the proposed objective face IQA model utilizing generative priors. As shown in Fig.~\ref{fig:arch}, the overall framework consists of the following three parts: a multistage encoder to map the target image into the latent GAN space, a \textcolor{black}{fixed} pretrained GAN model to generate intermediate reference features, and a quality predictor to obtain objective face quality estimations by fusing both target image features and intermediate reference features. Compared with conventional IQA models, which mostly employ a single encoder architecture for quality score regression, utilizing generative priors in the proposed framework has two advantages. First, by restricting target image features to the GAN latent space, semantically meaningful and attribute-aware representations can be encoded. Second, by feeding the latent codes forward, intermediate GAN-encoded statistics can be obtained and used as latent references for the challenging no-reference quality prediction task.

\subsection{Obtaining GAN Encoded Statistics}
\label{sec:4.1}

Due to the lack of inference ability of the GAN model, we first invert the target image $x$ into $N$ latent codes $w_1,w_2,\dots,w_N$ in the GAN input space. Specifically, we choose to train an encoder $E$ to map target images into the $\mathcal{W}^{+}$ space of StyleGAN2~\cite{karras2020analyzing}, a state-of-the-art GAN model capable of generating diverse facial images with high resolution and visual quality. 
Similar to~\cite{richardson2021encoding}, we encode $N$ style codes $w_i\in\mathbb{R}^{512}$ from multiple stages of a ResNet50~\cite{he2016deep} backbone, as follows:

\begin{equation}
    % \{\{w_i\}, f_E\} = E(x;\theta_{E}),\quad i=1,2,\dots,N
    (\mathbf{w}_N,f_E) = E(x;\theta_{E}),\quad \mathbf{w}_N=[w_1,w_2,\dots,w_N]
    \label{eq1}
\end{equation}

\noindent
where $f_E$ are intermediate features and $\theta_{E}$ are parameters of $E$. Latent codes $\mathbf{w}$ are then fed to the different scales of a fixed StyleGAN2 generator $G$ to produce a reconstructed result $\hat{x}_N$. During generation, we add $\mathbf{w}_N$ to the average latent code $\bar{\mathbf{w}}=[\bar{w_1},\bar{w_2},\dots,\bar{w_N}]$ in the pretrained $\mathcal{W}^{+}$ space to achieve a good initialization, as follows:

\begin{equation}
    \hat{x}_N = G(\mathbf{w}_N+\bar{\mathbf{w}}),
    \label{eq2}
\end{equation}

\noindent
where $\hat{x}_N$ denotes the reconstructed result from $x$.

To train the encoder, we optimize $\theta_E$ over the reconstruction error between $\hat{x}_N$ and $x$, as follows:

\begin{equation}
    \theta_{E}^{*}=\arg\min \mathcal{L}(\hat{x}_N, x),
    \label{eq3}
\end{equation}

\noindent
where $\mathcal{L}$ denotes the loss function. 

In this way, we train the encoder to map a target image $x$ into the GAN latent code space and obtain the intermediate features $f_E$ for quality prediction. However, to utilize rich generative priors, it is not enough to simply reconstruct the target image and extract the corresponding generative features. In the face IQA task, where target images are contaminated with distortions, the reconstructed results also contain degradation patterns and thus harm the GAN encoded statistics. To obtain facial statistics in the original GAN space, we take advantage of the interpretable and controllable attributes of the multiscale latent codes $\{w_i\}$. As latent codes at different scales are responsible for controlling level-specific facial attributes \cite{karras2019style, alharbi2020disentangled}, we observed that the low-level distortion attributes are inherently encoded in latent codes at finer scales, and the GAN statistics obtained in early stages are preserved. 
Therefore, during feed forward, we propose injecting the first $K, K < N$ codes into $G$ and discarding the last $N-K$ latent codes controlling the low-level details of $G$ to obtain generative representations $f_G^K$, which preserve the GAN encoded statistics.

\begin{equation}
    (\hat{x}_K, f_G^K) = G(\mathbf{w}_K+\bar{\mathbf{w}}), \quad \mathbf{w}_K=[w_1,w_2,\dots,w_K]
    \label{eq4}
\end{equation}

\noindent
where $\hat{x}_K$ is the reconstructed image with only the first $K$ codes injected, and $f_G^K$ denotes the intermediate generative features.

\begin{figure}[t]
\centering
\includegraphics[width=0.45\textwidth]{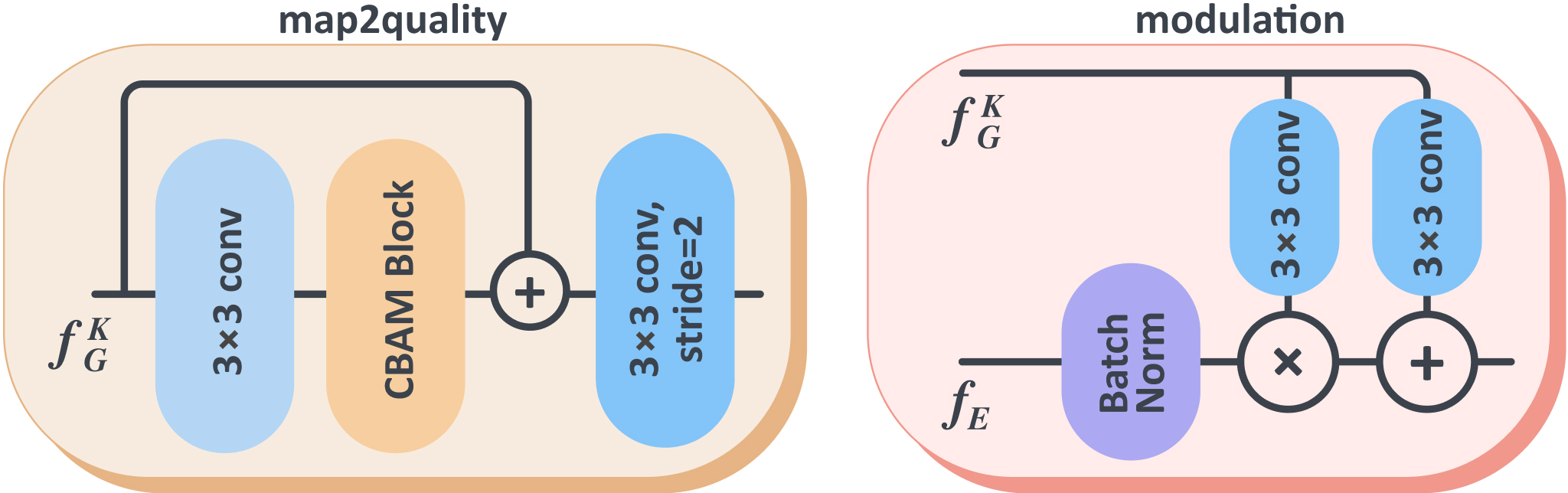}
\caption{The detailed architecture of map2quality and modulation blocks.
}
\label{fig:predictor}
\end{figure}

In this framework, the first $K$ codes are mainly responsible for reconstructing high-level facial attributes such as facial contours and organ shapes resembling the target image, and generative statistics are preserved since distortion patterns encoded in low-level codes are discarded. Notably, by directly discarding $N-K$ latent codes, we do not obtain reconstruction results that precisely match target images. However, in our IQA task, we do not need such perfect reconstruction results, and the results already obtain sufficient statistical priors to serve as latent references for solving the IQA problem. By further refining reference features to target image features, we combine them to make objective quality predictions.

\subsection{Quality Assessment with Generative References}

After obtaining the target image features $f_E$ and the generative reference features $f_G^K$, we refine and fuse them for quality prediction. Specifically, we extract high-level representations $f_E \in \mathbb{R}^{2048\times16\times16}$ from the last stage of $E$ to avoid the need for another encoding process and $f_G^K \in \mathbb{R}^{32\times256\times256}$ from the last scale of $G$ since it contains most of the generative information. We then apply an $1\times1$ convolution to $f_E$ and a series of map2quality blocks to $f_G^K$ for feature refinement. As mentioned in Section \ref{sec:4.1}, since the reconstructed structures do not perfectly match the target image, to refine the reference features, we use a CBAM \cite{woo2018cbam} with a residual connection inside each block to gradually adjust the features and a $3\times3$ convolution with stride 2 and doubled channel numbers to resize the features, as shown in Figure \ref{fig:predictor}. We then modulate $f_E$ by $f_G^K$, following the spatially adaptive denormalization operation proposed in \cite{park2019semantic}, as follows:

 \begin{equation}
    f_{mod} = \gamma^{n,c,y,x}(f_G^K)\frac{f_{E}^{n,c,y,x}-\mu _{E}^{c}}{\sigma _{E}^c}+\beta^{n,c,y,x}(f_G^K),
    \label{eq5}
\end{equation}

\noindent
where $\gamma^{n,c,y,x}(f_G^K)$ and $\beta^{n,c,y,x}(f_G^K)$ are elementwise modulation parameters after convolving $f_G^K$ with $3\times3$ kernels and $n,c,y,x$ are batch, channel and spatial indices, respectively. $\mu _{E}^{c}$ and $\sigma _{E}^{c}$ denote the channelwise mean and standard deviation values of $f_{E}$.

The operation modulates the distribution of target image features $f_E$ from its original distorted space to a generative reference space, thus serving refined reference features to the input. Finally, we concatenate $f_E$ with $f_{mod}$ and apply global average pooling followed by three fully connected layers to regress the features to the quality prediction score $\hat{q}(x)$.

\subsection{Objective Functions}

We use three loss functions, \emph{i.e.}, image reconstruction loss, regularization loss and quality prediction loss, to train our model. Image reconstruction loss ensures accurate GAN inversion results, which contain an $\mathcal{L}_{\rm{2}}$ loss, a perceptual loss $\mathcal{L}_{\rm{percep}}$ and a face identity loss $\mathcal{L}_{\rm{ID}}$, represented as follows:

\begin{equation}
\mathcal{L}_{\rm{2}}(x) = \|x-\hat{x}_N\|_2
\label{eq6}
\end{equation}

\begin{equation}
\mathcal{L}_{\rm{percep}}(x) = \|f_{\rm{percep}}(x)-f_{\rm{percep}}(\hat{x}_N)\|_2
\label{eq7}
\end{equation}

\begin{equation}
\mathcal{L}_{\rm{ID}}(x) = 1-\langle R(x), R(\hat{x}_N) \rangle,
\label{eq8}
\end{equation}

\noindent
where $f_{\rm{percep}}(\cdot)$ extracts perceptual features from a pretrained VGG \cite{simonyan2014very} model, and $R(\cdot)$ extracts identity vectors from a pretrained ArcFace \cite{deng2019arcface} model.

The regularization loss constrains encoder $E$ to output $\{w_i\}$ distributed within the latent generator space to avoid harming generative encoded statistics, as follows:

\begin{equation}
\mathcal{L}_{\rm{reg}}(x) = \|\{w_i\} - \bar{\mathbf{w}}\|_2 .
\label{eq9}
\end{equation}

The quality prediction loss further optimizes the parameters in the predictor, and we calculate the $\mathcal{L}_{\rm{1}}$ loss between the prediction result and subjective labels $q(x)$ as follows:

\begin{equation}
\mathcal{L}_{\rm{q}}(x) = |q(x)-\hat{q}(x)| .
\label{eq10}
\end{equation}

Finally, we sum the above loss functions with weights $\lambda_{i}, i = 1,2,\dots,5$ and jointly train the proposed model as follows:

\begin{equation}
\begin{split}
\mathcal{L}(x) = &\lambda_1\mathcal{L}_{\rm{2}}(x) + \lambda_2\mathcal{L}_{\rm{percep}}(x) + \lambda_3\mathcal{L}_{\rm{ID}}(x) \\
&+ \lambda_4\mathcal{L}_{\rm{reg}}(x) + \lambda_5\mathcal{L}_{\rm{q}}(x) .
\end{split}
\label{eq11}
\end{equation}

\section{Implementation Details}

In Table \ref{tab:arch}, we show the architectural details of the proposed model, including each module operation with its source input and output settings. The output size is shown in the order of $Channels\times Height\times Width$. It is worth noting that for a pretrained StyleGAN2 generating $512\times512$ resolution images, a total of 8 stages (1 stage without and 7 stages with upsampling) are included, and we combined every two stages in Table \ref{tab:arch} for simplicity.

\begin{table}[h]
\centering
\caption{Detailed architecture of our proposed model.}
\resizebox{0.48\textwidth}{!}{
\begin{tabular}{l|c|c|c}
\hline
\textbf{Module}                      & \textbf{Operation}   & \textbf{Input}                  & \textbf{Output Size}          \\ \hline
\multirow{8}{*}{Encoder}             & ResNet Stage1        & $3\times512\times512$ target image                               & 256 $\times$ 128 $\times$ 128 \\
                                     & ResNet Stage2        & ResNet Stage1                   & 512 $\times$ 64 $\times$ 64   \\
                                     & ResNet Stage3        & ResNet Stage2                   & 1024 $\times$ 32 $\times$ 32  \\
                                     & ResNet Stage4        & ResNet Stage3                   & 2048 $\times$ 16 $\times$ 16  \\ \cline{2-4} 
                                     & map2style1           & ResNet Stage1                   & 512 $\times$ 1 $\times$ 1     \\
                                     & map2style2           & ResNet Stage2                   & 512 $\times$ 1 $\times$ 1     \\
                                     & map2style3           & ResNet Stage3                   & 512 $\times$ 1 $\times$ 1     \\
                                     & map2style4           & ResNet Stage4                   & 512 $\times$ 1 $\times$ 1     \\ \hline
\multirow{4}{*}{Generator} & StyleGAN2 Stage1     & map2style4, $4\times4$ constant                           & 512 $\times$ 8 $\times$ 8     \\
                                     & StyleGAN2 Stage2     & map2style3, StyleGAN2 Stage1         & 512 $\times$ 32 $\times$ 32   \\
                                     & StyleGAN2 Stage3     & map2style2, StyleGAN2 Stage2        & 128 $\times$ 128 $\times$ 128 \\
                                     & StyleGAN2 Stage4     & map2style1, StyleGAN2 Stage3       & 32 $\times$ 512 $\times$ 512  \\ \hline
\multirow{7}{*}{Predictor}           & map2quality $\times$ 5 & StyleGAN2 Stage4                & 1024 $\times$ 16 $\times$ 16  \\
                                     & modulation           & ResNet Stage4, map2style        & 1024 $\times$ 16 $\times$ 16  \\
                                     & concat               & ResNet Stage4, modulation       & 2048 $\times$ 16 $\times$ 16  \\
                                     & global average pool  & concat                          & 2048                          \\ \cline{2-4} 
                                     & fully connection1    & global average pool             & 1024                          \\
                                     & fully connection2    & fully connection1               & 512                           \\
                                     & fully connection3    & fully connection2               & 1                             \\ \hline
\end{tabular}
}
\label{tab:arch}
\end{table}

We implemented our model with Pytorch, and StyleGAN2 is implemented based on its \href{https://github.com/rosinality/stylegan2-pytorch}{Pytorch version reimplementation}. The pretrained StyleGAN2 model is taken from \href{https://github.com/TencentARC/GFPGAN}{GFPGAN} \cite{wang2021towards}, where they provided the parameters for a $512\times512$ generator model. We selected $K=12$ for our model. During training, the batch size is set to 16, and the learning rate is set to $5\times10^{-5}$ and then decayed by a factor of 10 every 10 epochs. \textcolor{black}{During training, the StyleGAN2 decoder is fixed, and only the encoder and the predictor are optimized.} We trained the model with the Adam optimizer \cite{kingma2015adam} for a total of 25 epochs to report the final results. The whole model is trained using eight NVIDIA 1080Ti GPUs.

\section{Experiments}

\subsection{Setup}
We first randomly split the proposed GFIQA-20k dataset into a training subset (70\%, 14,000 images), a validation subset (10\%, 2,000 images) and a test subset (20\%, 4,000 images). For testing, we selected the best performing model with the highest SRCC on the validation set for performance comparisons. We use the SRCC, Pearson Linear Correlation Coefficient (PLCC), and Root Mean Square Error (RMSE) to evaluate the model prediction accuracy and monotonicity.

\subsection{How Does the Generic IQA Perform on the FaceIQA Task?}

To reveal the quality properties of face data, we first tested a cross database to observe how previous generic IQA datasets and models performed on the face IQA task. Specifically, we selected one synthetic IQA dataset LIVE \cite{sheikh:2006statistical} and four in-the-wild IQA datasets, including LIVE Challenge (LIVEC)~\cite{ghadiyaram2015massive}, KonIQ-10k~\cite{hosu2020koniq}, SPAQ~\cite{2020Perceptual} and KonIQ++~\cite{su2021koniq++}, for cross testing. We trained IQA models MEON~\cite{ma2018end}, HyperIQA~\cite{su2020blindly}, Koncept512~\cite{hosu2020koniq}, MT-A~\cite{2020Perceptual}, and BIQA~\cite{su2021koniq++} on the five datasets.
Among the testing models, MEON~\cite{ma2018end} and HyperIQA~\cite{su2020blindly} are state-of-the-art (SOTA) IQA methods that perform well on synthetic and authentic distortions, respectively, and the other models are proposed along with their training datasets. We tested the performance on the GFIQA-20k test subset and show the results in Table \ref{tab:corss test}.

\begin{table}[]
\centering
\caption{Performance comparisons by training on previous generic IQA datasets with a specified model and tested on the GFIQA-20k test subset.}
\begin{tabular}{lccc}
\hline
Dataset/Model        & SRCC$\uparrow$   & PLCC$\uparrow$   & RMSE$\downarrow$   \\ \hline
LIVE/MEON            & 0.6603 & 0.6371 & 0.1593 \\
LIVEC/HyperIQA       & 0.7501 & 0.7314 & 0.1055 \\
KonIQ-10k/Koncept512 & 0.8968 & 0.8925 & 0.0826 \\
SPAQ/MT-A            & 0.6980 & 0.7144 & 0.1282 \\
KonIQ++/BIQA         & 0.9225 & 0.9196 & 0.0720 \\ \hline
\end{tabular}
\label{tab:corss test}
\end{table}

From Table \ref{tab:corss test}, we observe that training on the synthetic IQA dataset LIVE did not give good predictions for in-the-wild face data. This result was foreseeable because of the domain gap between real world degradation and laboratory simulated distortions. The two authentic IQA datasets LIVEC and SPAQ also yielded relatively poor performances on the face data. This is probably because of the small number of training samples (1,162 images) contained in LIVEC and because of the bias in the smartphone photography images contained in SPAQ. Surprisingly, we found that KonIQ-10k and its extension KonIQ++ both performed relatively well (approximately 0.90 SRCC). The possible reason is that images from the KonIQ-10k dataset and the proposed GFIQA-20k dataset are both selected from YFCC100M \cite{thomee:2016}, resulting in a smaller domain gap. Despite this, there is still room for further performance improvement, and the following section discusses the development of face IQA models.

\subsection{Performance Evaluation with Competing Models}

In this subsection, we conduct performance comparisons of models trained with GFIQA-20k data. Due to the lack of baselines in the face IQA task, we first created diverse baseline models from different upstream tasks. Specifically, we selected ArcFace \cite{deng2019arcface} pretrained on the refined face recognition dataset MS1M \cite{guo2016ms}, Koncept512 \cite{hosu2020koniq} pretrained on the general IQA dataset KonIQ-10k \cite{hosu2020koniq}, and ResNet50 \cite{he2016deep} pretrained on the image classification dataset ImageNet \cite{deng2009imagenet}. We finetuned these models on the GFIQA-20k training subset and reported the results in Table \ref{tab:performance}. As seen, by simple transfer learning, all the baseline models achieved high performance (over 0.95 SRCC). The results demonstrate the effectiveness of the collected data in handling face IQA tasks.

\begin{figure*}[t]
\centering
\includegraphics[width=1\textwidth]{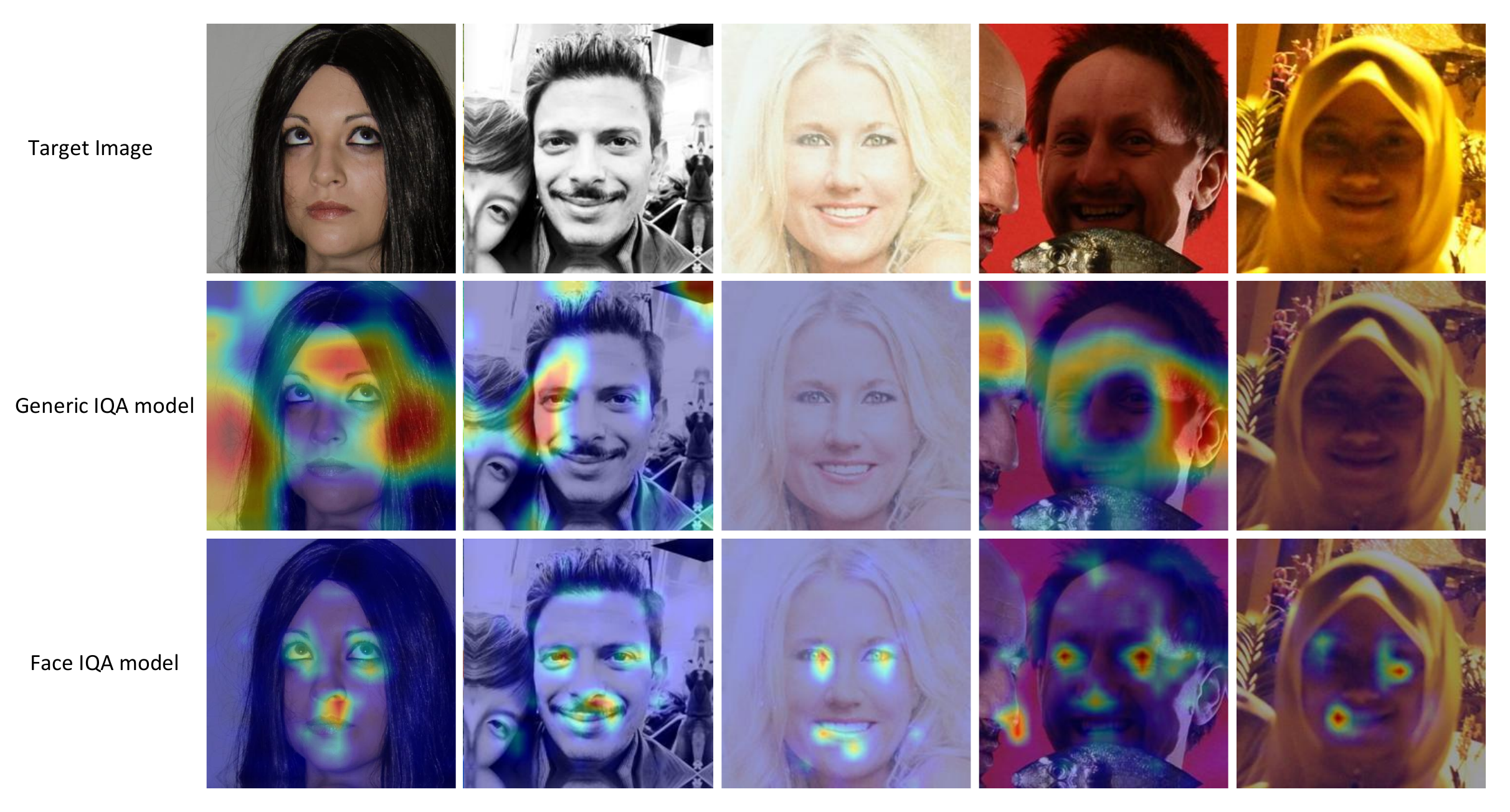}
\caption{\textcolor{black}{We compare heatmaps of the generic IQA model \cite{su2021koniq++} and face IQA model to visualize their perceptual differences regarding faces. It is clearly shown that the generic IQA model does not make quality predictions from faces, while the face model learns to consistently concentrate on perceptual critical regions such as eyes, noses and mouths to determine a face image's quality, regardless of various distortions.}}
\label{fig:comparisons}
\end{figure*}

\begin{figure*}[ht]
\centering
\includegraphics[width=1\textwidth]{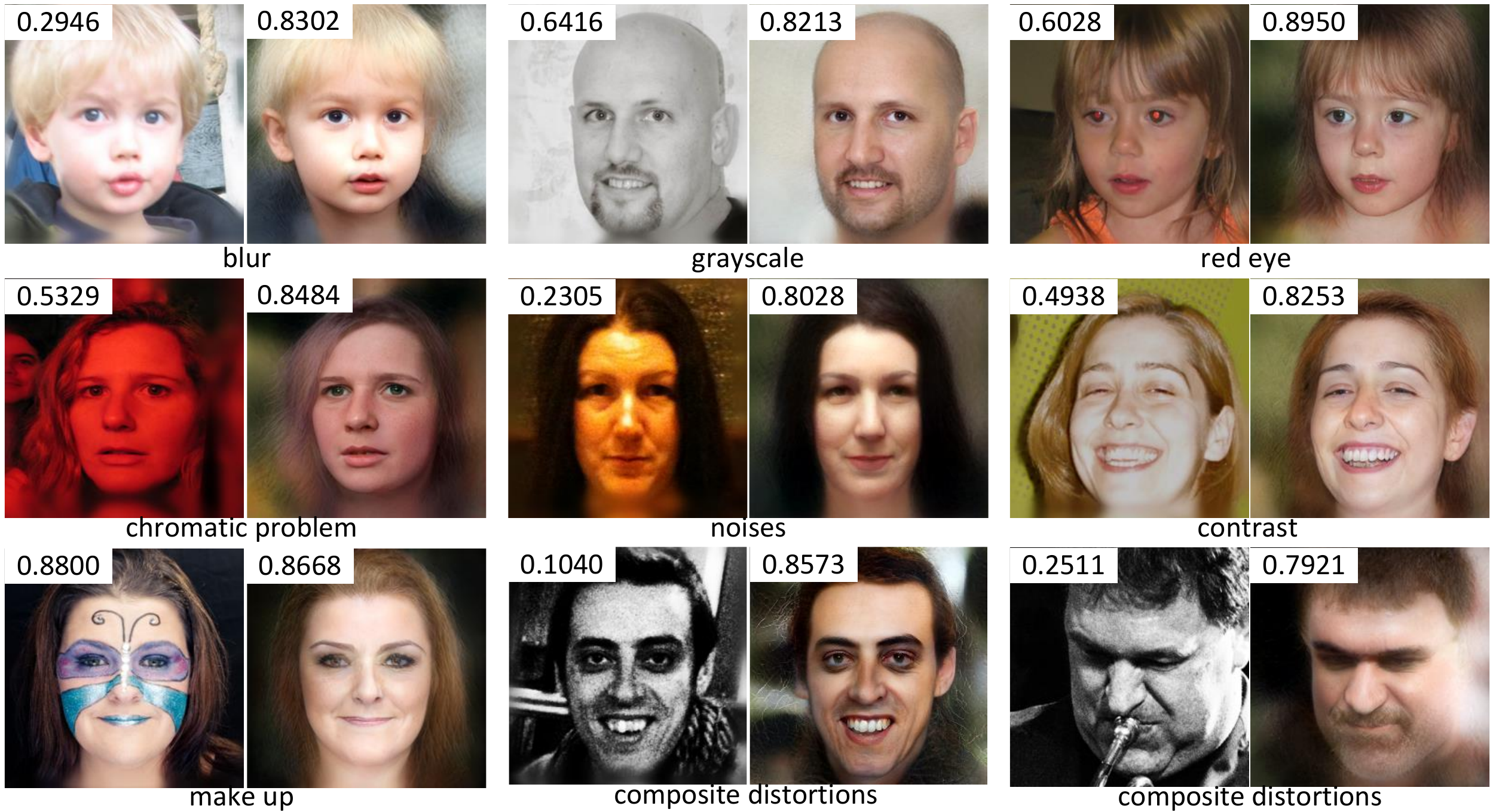}
\caption{\textcolor{black}{We visualize some generated latent reference images (right) with respect to their distorted images (left). Thanks to the powerful generative prior information, latent reference images are constructed despite various distortions contained in inputs and are able to facilitate the implementation of our quality prediction task. For each image, we also show the predicted quality score by MUSIQ \cite{ke2021musiq} for comparison.}}
\label{fig:examples}
\end{figure*}

\begin{table}[th]
\centering
\caption{\textcolor{black}{Performance comparisons of transfer learning of baseline models, SOTA IQA models and the proposed model.}}
\begin{tabular}{lccc}
\hline
Model      & SRCC$\uparrow$            & PLCC$\uparrow$            & RMSE$\downarrow$            \\ \hline
ArcFace    & 0.9505          & 0.9503          & 0.0588          \\
Koncept512 & 0.9520          & 0.9512          & 0.0572          \\
ResNet50   & 0.9629          & 0.9635          & 0.0504          \\ \hline
BRISQUE    & 0.7824          & 0.8055          & 0.1793          \\
CORNIA     & 0.8547          & 0.8616          & 0.1001          \\
HOSA    & 0.8861          & 0.8997          & 0.0945          \\
PQR       & 0.9519          & 0.9534          & 0.0551          \\
DBCNN   & 0.9609          & 0.9611          & 0.0520          \\
HyperIQA   & 0.9627          & 0.9635          & 0.0505          \\
MUSIQ      & 0.9630          & 0.9637          & 0.0503          \\ 
TRes & 0.9632          & 0.9638          & 0.0498          \\
\hline
Proposed   & \textbf{0.9643} & \textbf{0.9652} & \textbf{0.0486} \\ \hline
\end{tabular}
\label{tab:performance}
\end{table}

\textcolor{black}{We further compared the proposed model with the following eight general IQA models: BRISQUE~\cite{mittal2012no}, CORNIA~\cite{ye2012unsupervised}, HOSA \cite{xu2016blind}, PQR~\cite{zeng2017probabilistic}, DBCNN \cite{zhang2018blind}, HyperIQA~\cite{su2020blindly}, MUSIQ~\cite{ke2021musiq} and TRes \cite{golestaneh2022no}. Here, BRISQUE, CORNIA and HOSA are traditional IQA methods, PQR, DBCNN and HyperIQA are CNN-based deep learning models, and MUSIQ and TRes are transformer-based deep learning models. The selected deep learning models are all SOTA in-the-wild IQA models. Except for in the case of the traditional models, the training and testing runs are all repeated 10 times with random weight initialization for competing deep learning models, and the median results are reported in Table \ref{tab:performance}.
Among all the competing models, the proposed model outperformed the others on all three criteria. Statistical analysis also demonstrates the superior performance of the proposed model: by conducting a Student's t-test, the $p$ values between the proposed model and MUSIQ \cite{ke2021musiq} are 0.0100 for SRCC and 0.0032 for PLCC, and 0.0176 for SRCC and 0.0081 for PLCC against TRes \cite{golestaneh2022no}, where $p < 0.05$ indicates statistical significance.}

To further validate the effectiveness of the proposed model, in Table \ref{tab:amount}, we evaluated how the model performed with different training sample amounts. We compared the proposed model with the well-performing ResNet50 baseline and varied the training sample size from 10\% to 70\% of the images in the GFIQA-20k dataset, leaving the remaining images, except for the validation subset, for testing.
Similarly, the proposed model showed consistently superior prediction accuracy for variable training sample sizes.

\begin{table*}[]
\centering
\caption{Performance comparisons for different training sample amounts.}
\begin{tabular}{p{40pt}|p{45pt}|p{30pt}p{30pt}p{30pt}p{30pt}p{30pt}p{30pt}p{30pt}}
\hline
Criterion                         & Model    & 10\%            & 20\%            & 30\%            & 40\%            & 50\%            & 60\%            & 70\%            \\ \hline
\multirow{2}{*}{SRCC$\uparrow$}   & ResNet50 & 0.9480          & 0.9542          & 0.9581          & 0.9612          & 0.9626          & 0.9628          & 0.9629          \\
                                  & Proposed & \textbf{0.9484} & \textbf{0.9565} & \textbf{0.9609} & \textbf{0.9625} & \textbf{0.9632} & \textbf{0.9638} & \textbf{0.9639} \\ \hline
\multirow{2}{*}{PLCC$\uparrow$}   & ResNet50 & 0.9474          & 0.9537          & 0.9586          & 0.9618          & 0.9625          & 0.9632          & 0.9635          \\
                                  & Proposed & \textbf{0.9478} & \textbf{0.9570} & \textbf{0.9608} & \textbf{0.9623} & \textbf{0.9635} & \textbf{0.9643} & \textbf{0.9644} \\ \hline
\multirow{2}{*}{RMSE$\downarrow$} & ResNet50 & 0.0603          & 0.0550          & 0.0524          & 0.0521          & 0.0514          & 0.0507          & 0.0504          \\
                                  & Proposed & \textbf{0.0586} & \textbf{0.0538} & \textbf{0.0514} & \textbf{0.0503} & \textbf{0.0501} & \textbf{0.0489} & \textbf{0.0489} \\ \cline{1-9} 
\end{tabular}
\label{tab:amount}
\end{table*}

\subsection{Perceptual Comparison between Generic IQA and FaceIQA Data}

To understand the perceptual mechanism of deep models on face images and to reveal their difference from generic IQA, in this subsection, we visualize how models make their predictions when trained for generic IQA and for the GFIQA task. Specifically, we select the BIQA model from \cite{su2021koniq++} trained on the generic IQA dataset KonIQ++ and the proposed model trained on the GFIQA-20k dataset for comparison and draw their heatmaps \cite{zhou2016learning}
to understand how they perceive the face image quality. We show the results in Figure \ref{fig:comparisons} and make the following observations. First, although the generic IQA model \cite{su2021koniq++} performed relatively well on the GFIQA task in Table \ref{tab:corss test}, the predictions are not correctly made from the regions of the faces. It extracts features from different regions for different faces and thus might not be robust to diverse images. Second, as a comparison, when trained on the GFIQA-20k dataset, the model learns to capture critical face regions as quality representations, including eyes, noses and mouths. The result indicates the value and importance of the constructed dataset, which converges deep models on critical perceptual regions for the GFIQA task. Third, it is also interesting to find that although not explicitly constrained, the model learns on its own to consistently focus on the fixed regions when making quality predictions. We attribute the phenomenon to the subjective MOS being intrinsically perceptually biased toward these face organs, resulting in the model being optimized on these regions. The hypothesis further assists us in understanding some perceptual HVS mechanisms regarding faces. Since the deep model fits its perceptual mapping to the HVS, by examining how the model perceives face quality, we also assume that the HVS perceives face quality prior to perceiving facial organs. The finding is not only consistent with the neuroscience study that refrontal neurons in the HVS are selectively biased to recognize face identity \cite{o1997areal} but also unravels the precise and concrete regions to which the neuron cortex is particularly sensitive.

\subsection{Visualizing Generative References}
\label{sec5.E}

One of the benefits facilitated by utilizing generative priors is producing latent reference face images with preserved GAN statistics.
In this subsection, we visualize the reconstructed reference images to illustrate the effectiveness. \textcolor{black}{In Fig.~\ref{fig:examples}, we show pairs of distorted images $x$ and the reconstructed latent references $\hat{x}(K)$, as well as quality scores predicted by MUSIQ \cite{ke2021musiq} for both images. We include various in-the-wild distortions, including blur, color, contrast, noise, and composite distortions. Thanks to the rich prior information encoded in generative models, the reconstructed images are of high quality, \textit{i.e.}, mostly approximately 0.8-0.85 by MUSIQ scores, thus serving as latent references to the blind face IQA task. Since we impose loss constraints mainly on face regions, the generated latent reference images might not precisely match the target images in the hair or background regions. However, since the HVS is extremely sensitive to face regions such as eyes and mouths, the difference in hair or background regions does not contribute critically to the perceptual quality, which is also shown in the predicted reference image scores.}

\subsection{Ablation Study}

In this subsection, we conducted several ablation experiments to evaluate the effectiveness of the model design. 
We first compared our model configured with the baseline encoder ResNet50 trained by only $\mathcal{L}_{\rm{q}}$ loss. We then added StyleGAN2 and reconstruction loss to the model but did not fuse the latent reference features (w/o ref) to observe if encoding in generative latent space benefits model performance. Next, we evaluated how different values of $K$ affected the performance. We selected $K=4,8,12,16$ while keeping the other components fixed. Finally, we validated the designation of the quality predictor. We substituted the map2quality module with ordinary convolution blocks (w/o map) while the other parts remained fixed. We also removed the modulation block and simply concatenated $F_E$ and $f_G^K$ (w/o mod) to observe the effectiveness of the feature modulation block. The results are shown in Table \ref{tab:ablation}.

\begin{table}[ht]
\footnotesize
\centering
\caption{Ablation studies on different model configurations.}
\begin{tabular}{lccc}
\hline
model & SRCC$\uparrow$                       & PLCC$\uparrow$                       & RMSE$\downarrow$                       \\ \hline
baseline  & 0.9629                     & 0.9635                     & 0.0504                     \\
w/o ref   & 0.9629                     & 0.9636                     & 0.0504                     \\
K=4       & 0.9624                     & 0.9630                     & 0.0504                     \\
K=8       & 0.9627                     & 0.9635                     & 0.0505                     \\
K=12      & 0.9639                     & 0.9644                     & 0.0489                     \\
K=16      & 0.9637                     & 0.9644                     & 0.0492                     \\
w/o map   & 0.9631                     & 0.9635                     & 0.0503                     \\
w/o mod   & 0.9634                     & 0.9640                     & 0.0495 \\ \hline
full      & \textbf{0.9643}            & \textbf{0.9652}            & \textbf{0.0486}            \\ \hline
\end{tabular}
\label{tab:ablation}
\end{table}

\begin{figure*}[th]
\centering
\subfloat[MOS = 0.752, prediction = 0.647]{
\begin{minipage}[t]{0.48\linewidth}
   \includegraphics[width=0.48\textwidth]{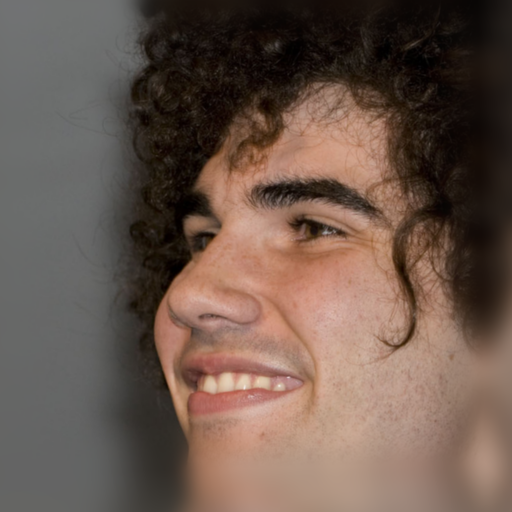}
   \includegraphics[width=0.48\textwidth]{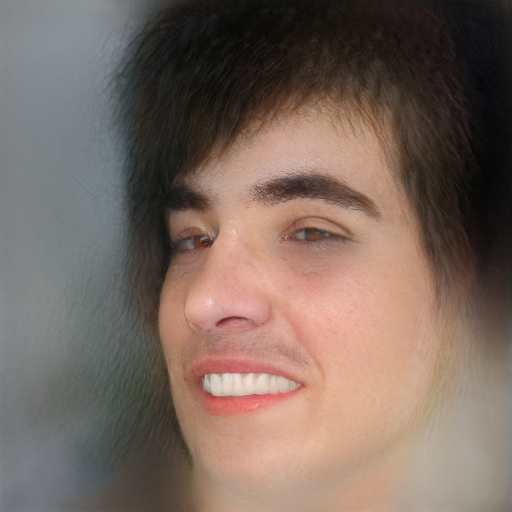}
\end{minipage}
}
\subfloat[MOS = 0.705, prediction = 0.571]{
\begin{minipage}[t]{0.48\linewidth}
   \includegraphics[width=0.48\textwidth]{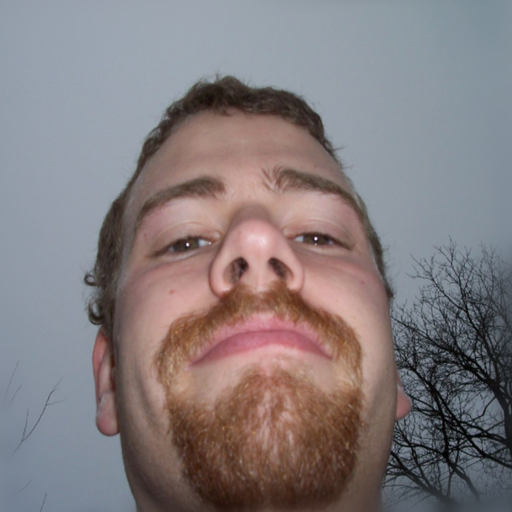}
   \includegraphics[width=0.48\textwidth]{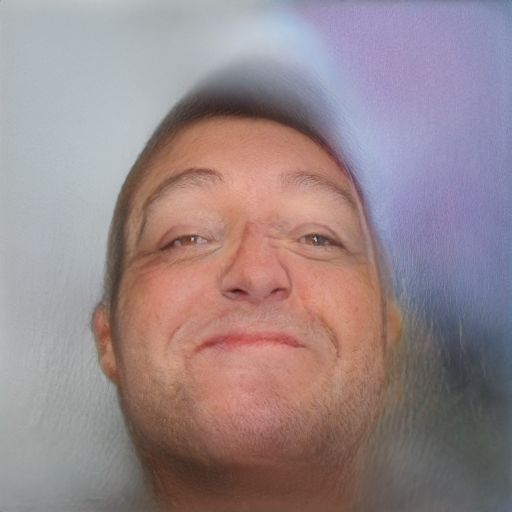}
\end{minipage}
}
\caption{\textcolor{black}{We show some image cases where quality prediction errors are larger than 0.1. The faces (left) are viewed at rotated angles, leading to inaccurate generative references (right) and, thus, inaccurate quality predictions. Including images with more diverse rotation angles to train more powerful generative priors might be a solution.}}
\label{fig:discussion}
\end{figure*}

From Table \ref{tab:ablation}, we make several observations. First, although not evident, encoding in generative latent space (w/o ref) slightly improved the model performance. Second, when extracting latent reference features from earlier generator stages (small $K$ values), the model showed inferior performance compared with the baseline. This is probably because, in the early stages, the generator is not able to encode enough statistics as a reference. However, when we extracted features from the latter stages, they outperformed the baseline model. Third, removing the map2quality or modulation module reduced the model performance, indicating the effectiveness of the proposed architecture of the quality predictor.

\subsection{\textcolor{black}{Complexity Analysis}}

\textcolor{black}{In this subsection, we compared the complexity of our model with that of two models, HyperIQA \cite{su2020blindly} and MUSIQ \cite{ke2021musiq}, in terms of computation complexity (FLOPs) and running time. The results are shown in Table \ref{tab:complexity}.}

\begin{table}[h]
\footnotesize
\centering
\caption{Complexity comparisons between the proposed model and other models.}
\begin{tabular}{l|ll}
\hline
Model    & FLOPs(G) & Time(s) \\ \hline
HyperIQA & 108.38         &  0.092       \\
MUSIQ    & 72.78         &   0.068 \\
Proposed & 240.75         &  0.205       \\ \hline
\end{tabular}
\label{tab:complexity}
\end{table}

\textcolor{black}{Although the computation complexity and running time of the proposed model are slightly larger than those of the two competitors due to the extra generative model, it achieves better prediction accuracy. Compared to other models, another benefit of introducing generative priors is that a totally training-free IQA model could be developed, as we will explain in Section~\ref{sec H}. }

\subsection{Developing a Total Training-free GFIQA Metric}
\label{sec H}

As shown in Section \ref{sec5.E}, thanks to generative priors, the model is able to produce images that are distributed close to the pristine image space as latent references. In this subsection, we further asked, with the latent references, are we able to develop a training-free GFIQA metric that fulfills the NR-IQA task without any training requirements? To answer this question, we modified the proposed model to a training-free model (proposed-TF) and evaluated its performance on the GFIQA-20k test set. Specifically, we extracted the generated image as a reference and calculated its LPIPS \cite{zhang2018unreasonable} distance to the target image as the quality measurement. Our underlying hypothesis is that since the generated images can serve as high-quality references, why not directly use FR-IQA models for quality prediction? Since both the StyleGAN2 and LPIPS models are off-the-shelf models, following the proposed framework, we are able to avoid the extra training process.

\begin{table}[ht]
\footnotesize
\centering
\caption{Performance comparison of the proposed training-free GFIQA metric with other opinion-unaware IQA metrics.}
\begin{tabular}{llll}
\hline
Model       & SRCC$\uparrow$   & PLCC$\uparrow$   & RMSE$\downarrow$   \\ \hline
NIQE        & 0.5549          & 0.5612          & 0.5824          \\
IL-NIQE     & 0.5806          & 0.5830          & 0.5538          \\
LPSI        & 0.2160          & 0.2504          & 0.9229          \\
QAC         & 0.2844          & 0.2932          & 0.8483          \\
dipIQ       & 0.4582          & 0.4847          & 0.6470          \\
RankIQA     & 0.5262          & 0.5435          & 0.4718          \\ \hline
Proposed-TF & \textbf{0.7012} & \textbf{0.7236} & \textbf{0.3407} \\ \hline
\end{tabular}
\label{tab:tf}
\end{table}

We compare the proposed training-free metric with other opinion-unaware IQA models, including NIQE \cite{mittal2012making}, IL-NIQE \cite{zhang2015feature}, LPSI \cite{wu2015highly}, QAC \cite{xue2013learning}, dipIQ \cite{ma2017dipiq} and RankIQA \cite{liu2017rankiqa}. Among the compared methods, NIQE \cite{mittal2012making}, IL-NIQE \cite{zhang2015feature} and LPSI \cite{wu2015highly} are totally blind IQA estimators, and the rest are trained on pseudo image quality labels. We show the results in Table \ref{tab:tf}, and we make the following observations. First, as shown, the proposed metric outperformed all the competitors by a large margin on the GFIQA-20k test set. Since the competing models mainly focus on the synthetic IQA task, the proposed model showed its superior effectiveness on the more challenging in-the-wild IQA task. Second, we also found that the two deep learning-based models dipIQ \cite{ma2017dipiq} and RankIQA \cite{liu2017rankiqa} actually performed worse than the totally training-free methods NIQE \cite{mittal2012making} and IL-NIQE \cite{zhang2015feature}. The result indicates that the potential risk of overfitting exists in the two deep learning-based models, which are trained with images containing synthetic distortions. Third, compared with the training-based model, the training-free framework still performed worse. This is probably due to the discrepant perceptual mapping learned by LPIPS to the face perceptual domain and the imperfect generative references of some
flawed face reconstruction cases, as will be explained in Section. \ref{sec6}. Nevertheless, training-free models are commonly agreed to be more robust to unseen data \cite{kancharla2021completely}\cite{li2022blindly}; thus, we expect the proposed training-free IQA framework to be applied to challenging real-world IQA applications.

\begin{figure*}[ht]
\centering
\begin{minipage}[b]{0.24\linewidth}
\centering
\includegraphics[width=\textwidth]{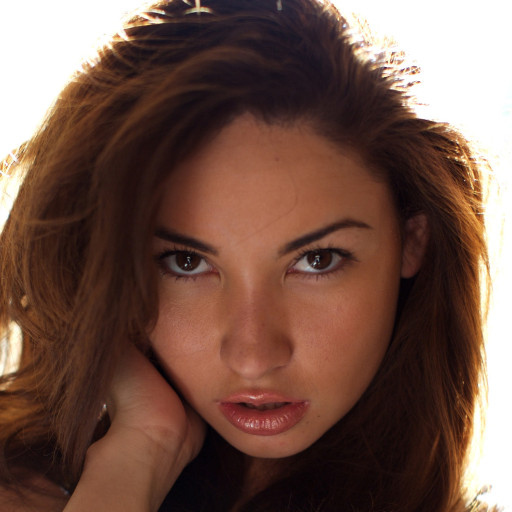}
%\caption{default}
%\label{fig:figure1}
\centerline{(a)}
\end{minipage}
%\hspace{0.5cm}
\begin{minipage}[b]{0.24\linewidth}
\centering
\includegraphics[width=\textwidth]{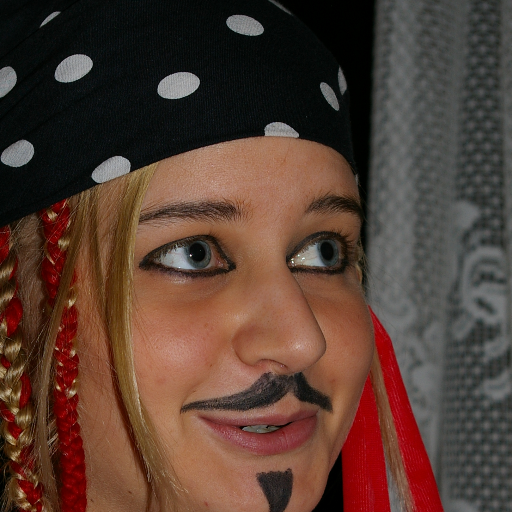}
\centerline{(b)}
%\caption{default}
%\label{fig:figure2}
\end{minipage}
\begin{minipage}[b]{0.24\linewidth}
\centering
\includegraphics[width=\textwidth]{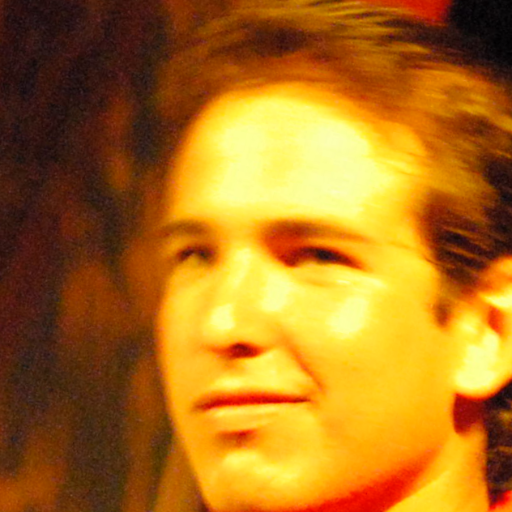}
%\caption{default}
%\label{fig:figure1}
\centerline{(c)}
\end{minipage}
%\hspace{0.5cm}
\begin{minipage}[b]{0.24\linewidth}
\centering
\includegraphics[width=\textwidth]{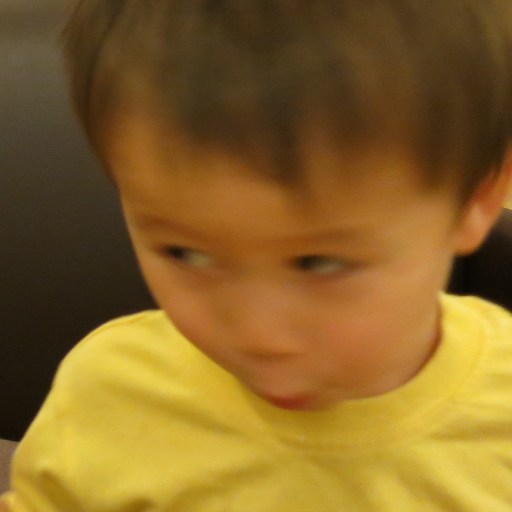}
\centerline{(d)}
%\caption{default}
%\label{fig:figure2}
\end{minipage}
\caption{We show face examples with various technical qualities and attractiveness. (a) High quality (MOS = 0.910) and high attractiveness, (b) High quality (MOS = 0.901) but low attractiveness, (c) Low quality (MOS = 0.130) but high attractiveness, and (d) Low quality (MOS = 0.138) and low attractiveness.} \label{fig:attract}
\end{figure*}

\section{Discussion}
\label{sec6}

\textcolor{black}{
Although the proposed model showed its superiority in face quality prediction, we find some limitations regarding the generative priors. By selecting test images with prediction errors greater than 0.1 (10\% of the quality score range), we find that images with rotated faces usually lead to unsatisfactory reconstructed results and poorer quality predictions, as shown in Fig.~\ref{fig:discussion}. We assume this is because the generative prior model StyleGAN2 was mostly trained with frontally viewed samples while few images of rotated faces were included. Thus, the generator can produce frontally viewed faces but underperforms otherwise. To address this issue, training the generative model with faces viewed from diverse view angles to provide more powerful priors might be a solution, and we leave the task for future work. It is also worth noting that in our proposed framework, the prior model could be substituted by others; therefore, with the development of more powerful generative models, the proposed IQA model should also benefit and perform better, which we expect to observe in the future.
}

We also clarify that technical face image quality should not be confused with face attractiveness, \emph{i.e.}, aesthetic. In the subjective study, apart from giving a precise definition of technical image quality in the instruction, we provided a few examples to teach freelancers how to differentiate them. Fig.~\ref{fig:attract} shows some face examples in GFIQA-20k with various qualities and attractiveness. It shows the MOSes are consistent with technical quality rather than attractiveness, which demonstrates the effect of face attractiveness was minimized in the proposed dataset.

\section{Conclusion}

We created GFIQA-20k, the largest annotated in-the-wild database for face image quality prediction. The dataset contains 20,000 faces of diverse individuals in various circumstances. Furthermore, to accurately predict face image quality, we introduce generative prior information to the IQA task for the first time. The proposed model makes use of rich statistics encoded in pretrained deep generative models. Our experiments validated its superiority relative to existing works. We expect both the dataset and the model will be valuable in face processing and general IQA research.

\bibliographystyle{IEEEtran}

\bibliography{TMM-final-version}

\end{CJK}

\end{document}